\documentclass[10pt, conference, compsocconf]{IEEEtran}
\usepackage{subfigure}
\usepackage{blindtext}
\usepackage{graphicx}
\usepackage{cite}
\usepackage{csquotes}
\usepackage{caption}
\usepackage{amssymb}
\usepackage{amsmath}
\usepackage{siunitx}
\usepackage{multicol}
\DeclareCaptionLabelSeparator{periodspace}{.\quad}
\ifCLASSINFOpdf
\else
\fi
\begin{document}
%
\title{Deep Learning Inference on Embedded Devices: Fixed-Point vs Posit}


\author{\IEEEauthorblockN{Seyed H. F. Langroudi, Tej Pandit, Dhireesha Kudithipudi}
\IEEEauthorblockA{Neuromorphic AI Lab, Rochester Institute of Technology\\
Rochester, New York 14623\\
\{sf3052,tnp5438,dxkeec\}@rit.edu}
}


%


\maketitle

\begin{abstract}
Performing the inference step of deep learning in resource constrained environments, such as embedded devices, is challenging. Success requires optimization at both software and hardware levels. Low precision arithmetic and specifically low precision fixed-point number systems have become the standard for performing deep learning inference. However, representing non-uniform data and distributed parameters (e.g. weights) by using uniformly distributed fixed-point values is still a major drawback when using this number system. Recently, the posit number system was proposed, which represents numbers in a non-uniform manner. Therefore, in this paper we are motivated to explore using the posit number system to represent the weights of Deep Convolutional Neural Networks. However, we do not apply any quantization techniques and hence the network weights do not require re-training. The results of this exploration show that using the posit number system outperformed the fixed point number system in terms of accuracy and memory utilization.  

\end{abstract}


%
\IEEEpeerreviewmaketitle

\section{Introduction}
Deep learning, as a particular form of hierarchical representational learning \cite{Alecun2015deep}, has shown promise in several applications such as computer vision \cite{DensNetCVPR}, natural language processing \cite{GNMT}, speech recognition \cite{deepspeechu2}, robotics \cite{deepRobotic2017} and medical applications \cite{skin}. The success of deep learning stems from its ability to learn from raw and unstructured data \cite{Alecun2015deep}. Deep Convolutional Neural Networks (DCNN) are commonly used in Deep learning, with stochastic gradient descent as their learning mechanism \cite{DeepLearningbook}. 

Although DCNNs achieve state-of-the-art accuracy as compared to other machine learning approaches, they exhibit shortcomings such as long latency, power inefficiency, and long training durations. For instance, training ResNet-50  (50 layers) \cite{ResNet2016} on the ImageNet dataset \cite{ImageNet2015} requires 256 GPUs \cite{ResNet-50-1hours}. Another example, AlphaGo, was trained for months with 1202 CPUs and 176 GPUs to beat Lee Sedol, an 18-time world champion, at the strategic board game "Go"\cite{masteringAlphago}. According to these examples, training deep neural networks even with the resources in data centers has a lot of limitations. On the other hand, deep learning inference is less complex than deep learning training. Moreover, the limitations for implementing deep learning inference on conventional hardware such as CPUs and GPUs has been addressed by digital neuromorphic chips such as TPU \cite{TPU2017}. However, this chip is specially designed for data centers. Therefore, designs for digital neuromorphic chips to implement DCNNs with real-time performance on low-power embedded platforms, mobile devices, and IoT devices, are currently in the research exploration phase \cite{deeprebirth, IOTdeep}. Low precision arithmetic is a common approach to reduce power consumption and improve the real-time performance of deep learning applications on embedded devices.

Among different number systems used for performing deep learning inference with low precision arithmetic, the fixed-point number system shows the most promising trade-off between accuracy and computational complexity \cite{judd2016proteus,Gysel2016,Quntized,mishra2017apprentice}. However, real numbers are represented uniformly by a fixed-point number system which is not suitable for deep learning applications since the weights and data have a non-uniform distribution \cite{Gysel2016}. Recently, a posit number system was proposed as an alternative to the floating point number system \cite{gustafson2017beating}. This number system has a unique non-linear numerical representation characteristic for all numbers in a dynamic range which distinguishes it from other number systems such as fixed and floating point. As a result, in this introductory paper, we are motivated to explore the use of the posit representation in DCNNs for digit recognition and image classification tasks. 

We compare the fixed-point number system and the posit number system to represent weights of three DCNNs with 4, 5 and 8 layers on MNIST \cite{lecun1998gradient}, Cifar-10 \cite{Cifar-10stateart} and ImageNet \cite{ImageNet2015} datasets respectively. The posit number system outperformed the fixed-point number system in terms of accuracy and memory utilization when the two number systems are compared under the constraint that they both have the same dynamic range ([-1,1]).

The rest of this paper presents previous works on low precision deep learning inference and introduces the posit number system in section \ref{sec:Background}, the proposed DCNNs using the posit number system are discussed in section \ref{sec:DCNNs with posit number system}, the comparative results, in terms of accuracy and memory utilization, between the posit number system and the fixed-point number system to represent weights of proposed DCNNs are presented in section \ref{sec:evalution}. 
 
\section{Background}
\label{sec:Background}

\subsection{Low precision deep learning inference}
\label{Low precision}
In recent literature studies on DCNNs, focusing on improving the computational efficiency during inference by using limited precision for weights and activations, Judd et al. represent weights by the dynamic fixed-point number system and perform computations using the floating point number system \cite{judd2015reduced}. In this approach, the energy consumption for memory access operations obtained while implementing different deep neural networks on various datasets is reduced by an average of 15\% \cite{judd2015reduced}. Following this research, the 8-bit floating point number system used to represent weights of AlexNet and VGG-16 \cite{VGG162014} and was evaluated on the ImageNet dataset \cite{ReducedPrecisionMemory}. The results indicated that it is possible to represent 20\% of the weights in the 8-bit floating point representation with less than 1\% degradation of accuracy. Finally, Gysel et al. successfully performed deep learning inference using the AlexNet architecture on the ImageNet dataset, with 8-bit dynamic fixed-point weights and 8-bit dynamic fixed-point data, resulting in less than 1\% degradation of accuracy \cite{Gysel2016}. However, the networks needed to be retrained in order to attain this level of accuracy. 

After the success of performing deep learning inference by using an 8-bit precision representation of weights and data, researchers have been further motivated to squeeze the representation to below 8-bits, in particular, the 1-bit (binarized representation) \cite{Binarized2016,Quntized} and 2-bit (ternarized representation) \cite{terneryweight2016,mishra2017apprentice}. Although, by using these representations, multiplication operations in a deep neural network are removed or converted to sign detection operations, the corresponding significant degree of degradation in accuracy overwhelms the computational advantage. Therefore, evaluating a deep learning inference model with 8 layers or more (e.g. AlexNet, GoogLeNet) on large datasets (e.g., ImageNet), with less than 8 bits to represent each of the weights and data values, without substantial accuracy degradation and/or retraining, is still an open question. 
     
\subsection{Posit number system}
The posit number system is a type of tapered accuracy number system \cite{morris1971tapered}, which means that numbers with small exponents are more accurate than numbers with large exponents \cite{gustafson2017beating}. The challenges encountered by the floating-point number system such as manipulating overflow, underflow, double zero and exception are addressed in this number system. The posit number system format defined as $P_{(n, es)}$ where \emph{n} refers to the total number of bits in this system and \emph{es} indicates the number of exponent bits \cite{gustafson2017beating}. Each number in this system, as shown by Eq. \ref{equ:equ2} \cite{gustafson2017beating}, is indicated by $useed$, $exponent$, $r_{value}$ (responsible for finding the number regime) and $fraction$ (to indicate the precision).
\begin{equation}
\label{equ:equ2}
\resizebox{.43 \textwidth}{!} 
{
	$X=(-1)^{sign} \times (useed)^{r_{value}} \times 2^{exponent} \times (1+fraction)  $
}    
\end{equation}

\begin{figure}[h!]
\begin{center}
   \includegraphics[width=1\linewidth]{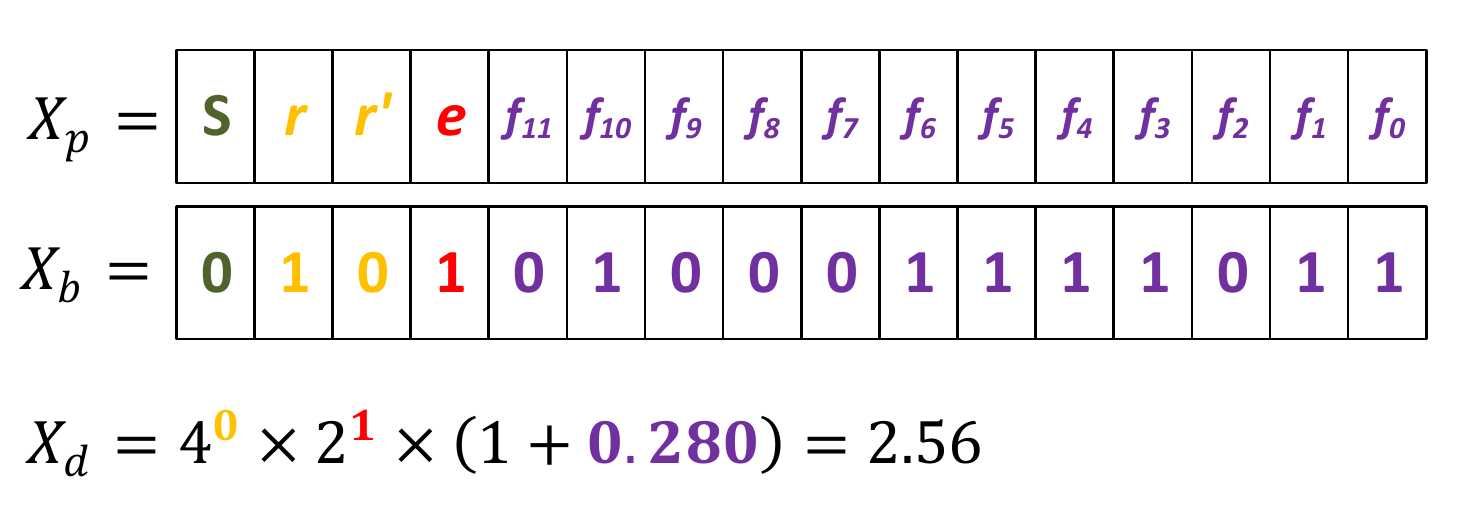}
\end{center}
 \caption{Representation of a number in the posit number system with $P_{(16,1)}$ format  \cite{gustafson2017beating}}
\label{fig:PositRepresentation}
\end{figure}

For instance, 2.56 in the posit number system with $P_{(16,1)}$ format is represented by 4 as a $useed$, 1 as an $exponent$, 0 as an $r_{value}$ and 0.280 as a $fraction$ as shown by Fig. \ref{fig:PositRepresentation}. Note that the conversion from decimal floating point to posit numbers are explained in detail in section \ref{sec:DCNNs with posit number system}.

\section{DCNNs with posit representation}
\label{sec:DCNNs with posit number system}
\begin{figure*}[ht]
\begin{center}
   \includegraphics[width=0.75\linewidth]{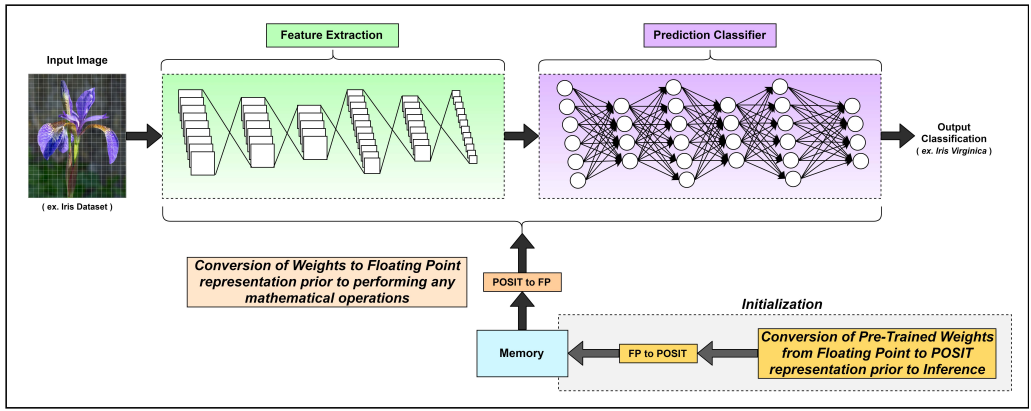}
\end{center}
 \caption{High level overview of the DCNN architecture implementation which uses the posit number system to represent the weights}
\label{fig:DCNNs}
\end{figure*}
In this paper, we explore the effects of using the posit number system (used to represent weights), on the accuracy and memory utilization of the DCNN during inference. To achieve this goal, the weights are converted from the original single floating number system to the new posit number system during memory read and write operations. On the other hand, the posit number system is converted back to a single floating point number system during computational operations as needed by standard computing architectures. The proposed DCNNs architecture is shown in Fig. \ref{fig:DCNNs}. This architecture is similar to DCNNs architecture which is proposed by \cite{judd2015reduced} except we use a posit number system which has advantages to represent weights of DCNNs non-uniformly. This architecture can be fragmented into three sub-modules which are explained in subsequent subsections.   

\subsection{Conversion from posit to floating point}
The first step is to convert the posit number to a decimal floating point number and then convert to a binary floating point number. The conversion from a posit to a decimal floating point number is divided in four steps \cite{gustafson2017beating}: (i) extracting the sign bit; (ii) extracting the regime bit; (iii) extracting the exponent bit; (iv) extracting the fraction bit. The most significant bit in posit representation indicates the sign bit. The regime bit is represented by unary arithmetic \cite{gustafson2017beating}. Therefore, when extracting the regime value ($r_{value}$), the algorithm starts to count the number of consecutive one's or zero's after a sign bit until it reaches a bit of the opposite value (zero or one respectively). Then, the result is negated if the bits counted are zeros or is decremented by 1 if the bits counted are ones. The exponent bits are represented by an unsigned integer and are thus easily extracted from the posit number bit string. The rest of bits in the bit string are fraction bits. The decimal floating point number is converted into a binary floating point number by dividing or multiplying by 2 until the number is in the range of $[1, 2)$\cite{gustafson2017beating}. 
\subsection{DCNNs}
In DCNNs, the features are extracted using convolutional layers. This feature is shown by vector $F_{n}= (f_{1}, f_{2}, ... , f_{m})$, where $n$ indicates the number of images in the dataset, and $m$ shows the feature vector dimension. Then the features are classified by a network of fully connected layers. In the last layer, the softmax layer is used as a classifier to minimize $ y - f^*(x,w)$; where y defines a label, x denotes an input to the softmax layer, w denotes the weights in softmax layer and $f^*$ is the best approximation function \cite{DeepLearningbook}.

\subsection{Conversion from floating point to posit}
This conversion consists of two steps \cite{gustafson2017beating}: (i) converting the binary floating point number system to the decimal floating point number system; (ii) converting the binary floating point number system to the posit number system. The first conversion is performed by multiplying the fraction by two raised to the power of the exponent. Then the decimal floating point number achieved in the first step is converted into a posit number by dividing or multiplying it by two until the number is in the range $[1,useed)$ in order to find the regime bit. Then, this process is continued until the number is in the range $[1,2)$ in order to find the exponent. The remaining bits are the fraction \cite{gustafson2017beating}.

\section{Evaluation}
\label{sec:evalution}
The new approach is evaluated on three datasets: (i) MNIST dataset; (ii) CIFAR-10 dataset; (iii) subset of the ImageNet dataset. The MNIST dataset (handwritten numerical digits dataset), and others datasets are collected for assessing the performance of new techniques on basic image recognition tasks. Different DCNNs are used for each dataset, and the single floating point number system is selected for a baseline implementation. The baseline is implemented using the Keras API \cite{keras} and the accuracy results are as shown in Table \ref{table:1-1}.

\begin{table}[ht!]
\caption{Top-1 accuracy of 3 different neural networks}
\centering
\resizebox{0.48\textwidth}{!}{
\begin{tabular}{|c|c|c|c|c|c|} 
\hline
    Task & Dataset & \# inference set & Network & layers & Top-1 accuracy \\
    \hline \hline
    Digit classification & MNIST & 10000 & LeNet & 2 Conv and 2 FC & 99.03\% \\
    \hline
    Image classification & CIFAR-10 & 10000 & Convnet & 3 Conv and 2 FC & 68.45\% \\
    \hline
    Image classification & ImageNet & 10000 & AlexNet & 5 Conv and 3 FC &    55.45\% \\
    \hline
\end{tabular}
}
\label{table:1-1}
\end{table}

\begin{figure*}[h!tb]
\centering
\subfigure[]{\includegraphics[width=50mm, height=40mm]{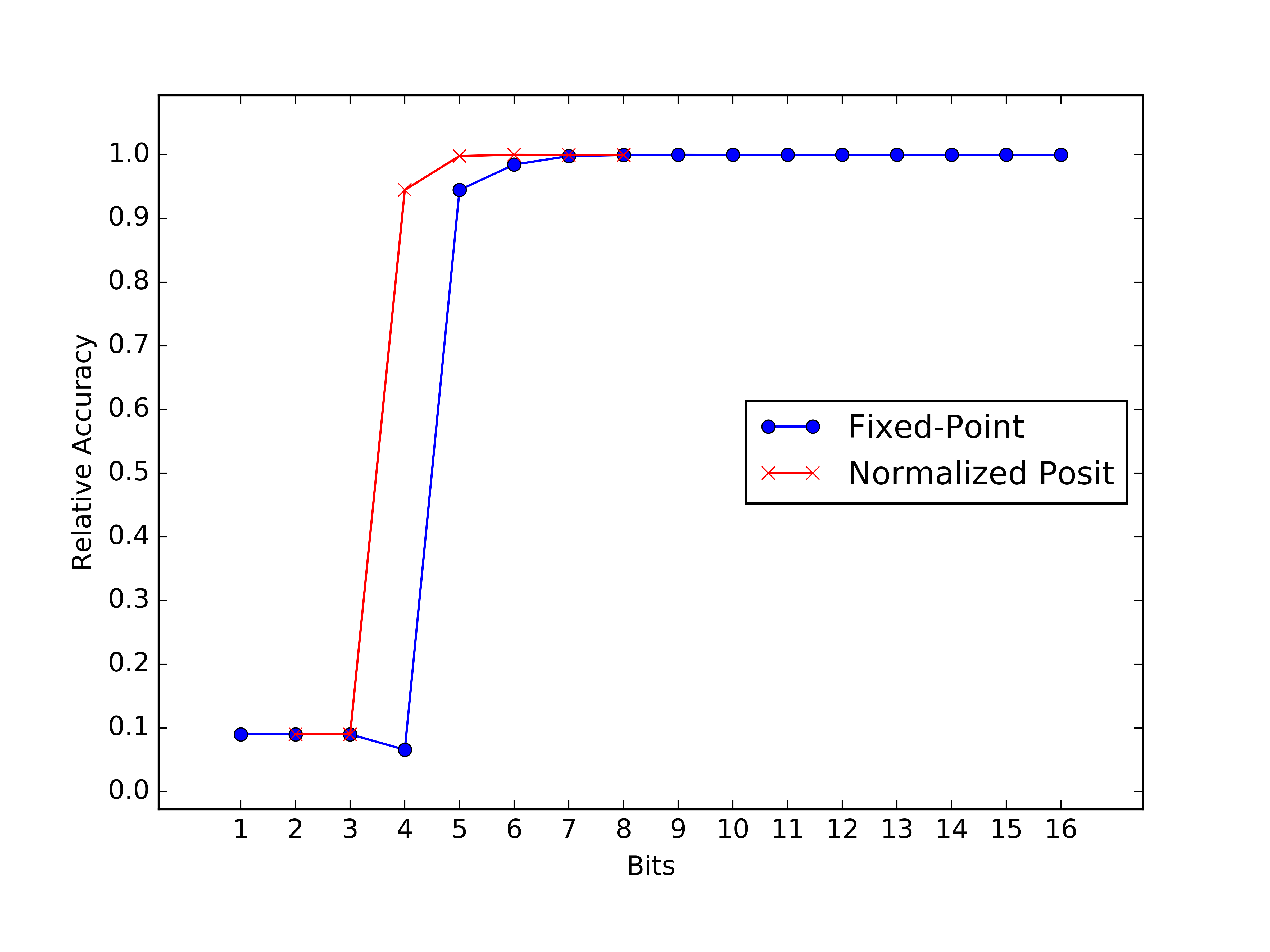}}
\subfigure[]{\includegraphics[width=50mm, height=40mm]{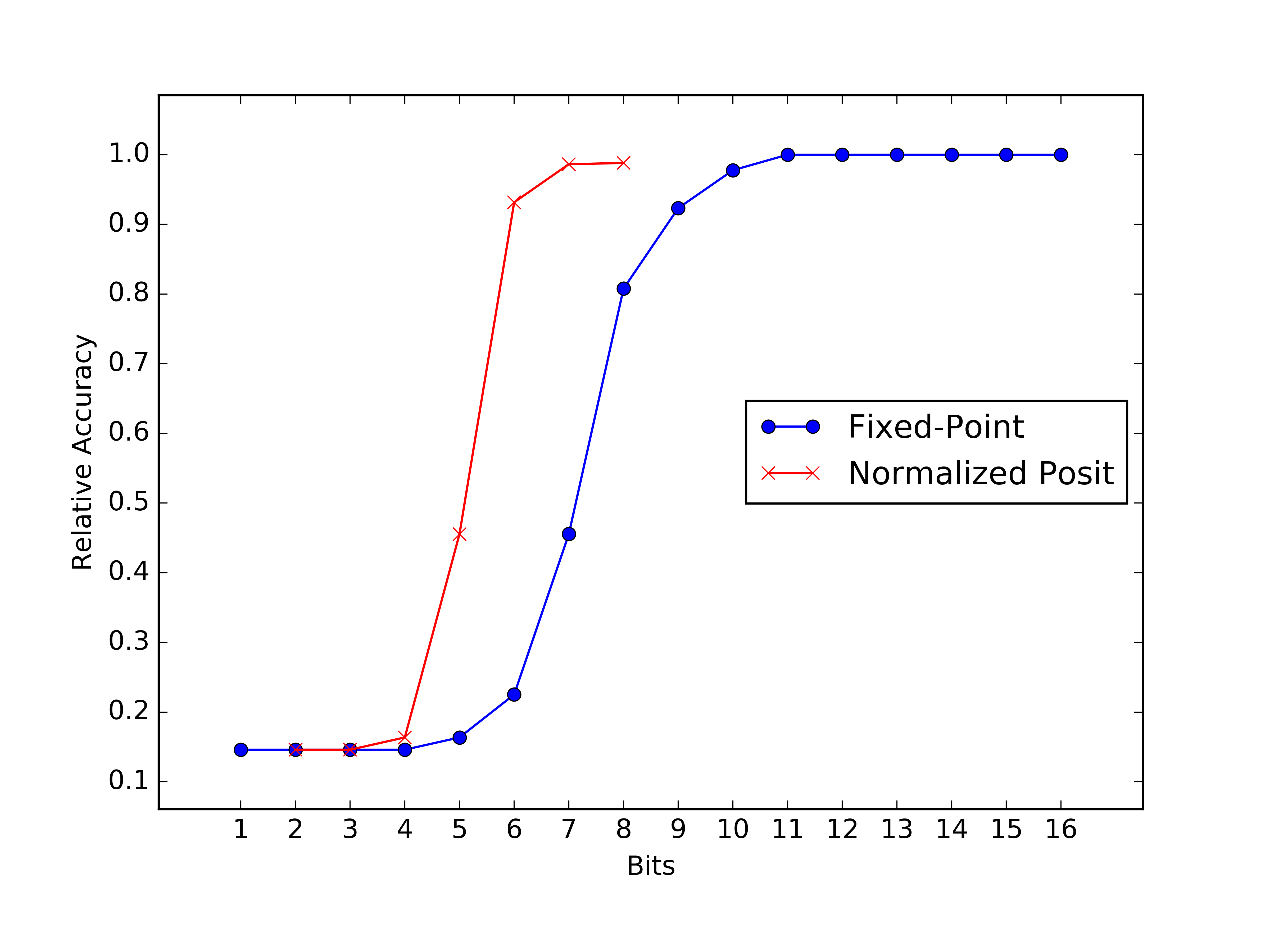}}
\subfigure[]{\includegraphics[width=50mm, height=40mm]{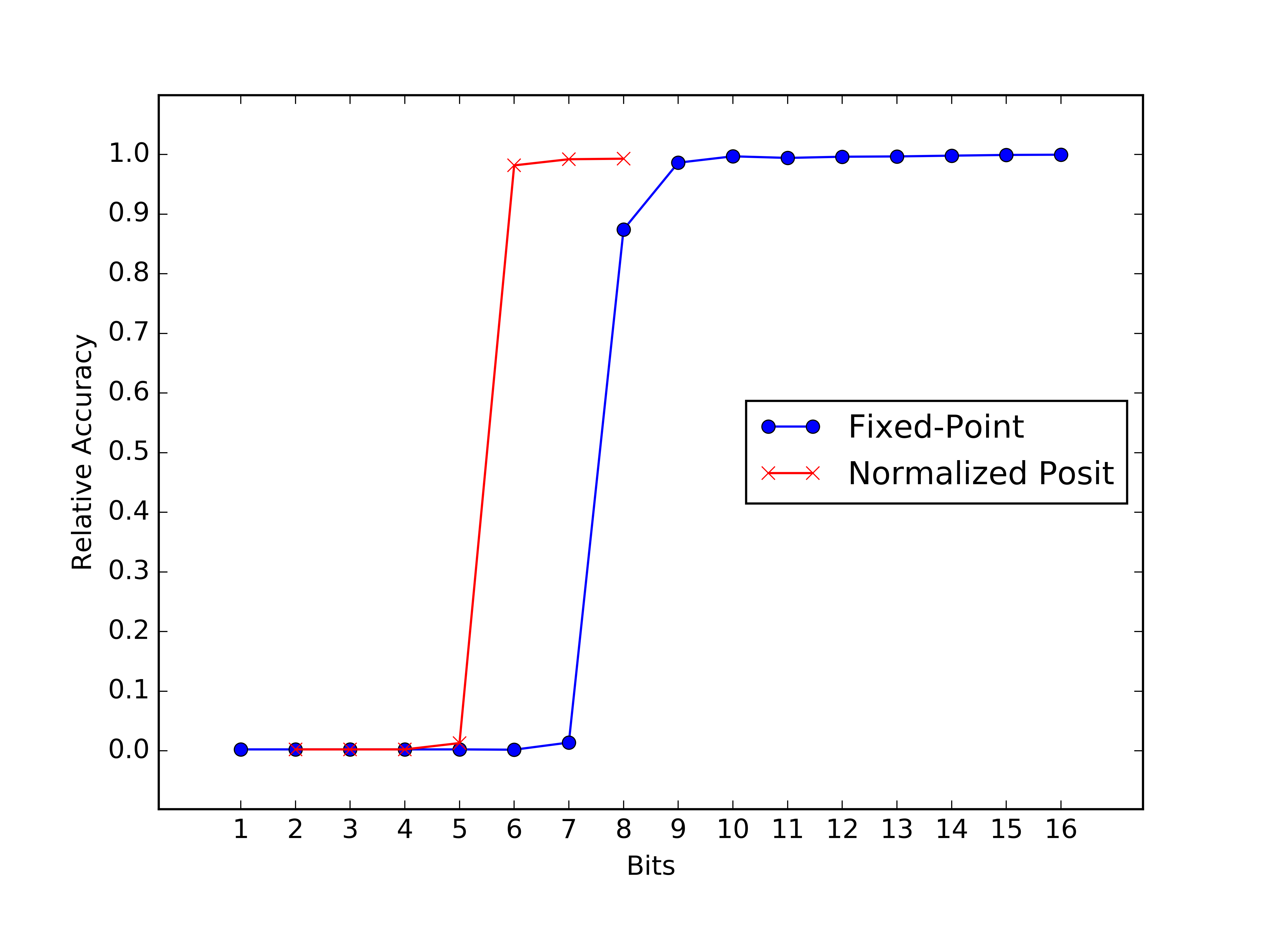}}
\caption{Results showing the relative accuracy to the baseline for DCNN implementations on various datasets with representation of weights using variable length fixed point and normalized posit number systems.
(a) Relative Accuracy results for LeNet on MNIST dataset. (b) Relative Accuracy results for ConvNet on Cifar-10 dataset. (c) Relative Accuracy results for AlexNet on ImageNet dataset.}
\label{fig:results}
\end{figure*}


In this paper, the weights are represented by a variable length fixed-point number system (with a maximum bit length of 16 bits) and 8-bit posit number system. To represent weights in the variable fixed-point number system only one bit is considered for the integer part and the fractional part is varied in a range of [0,15] bits, as most of the weights in well explored DCNNs are in the $[-1,1]$ interval. To represent weights with the posit number system, we selected the $P_{(i,0)}$ format where $i$ is varied within the range [2,8]. Note that the exponent selected is zero. The reason behind this selection is because the dynamic range of the posit number system with a zero exponent is the closest approximation to the weights' dynamic range, as compared to other possible options for exponent value. Among these posit formats, $P_{(2,0)}$ has the smallest dynamic range ($[-1,1]$), while other posit formats have a larger dynamic range. However, variable length fixed point number systems have the same dynamic range. Therefore, we are motivated to normalize all the formats in posit number system and call it the normalized posit number system. In this version of the posit number system, all of the formats have the same dynamic range of $[-1,1]$. The relative accuracy results for different tasks are shown in Fig. \ref{fig:results}.  


The normalized posit number system outperformed other number systems in terms of accuracy with fewer bits. The results demonstrate that it is possible to perform LeNet, ConvNet, and AlexNet with 5 bits, 7 bits and 7 bits respectively, using the posit number system, with less than 1\% accuracy degradation in comparison to performance of the same networks with 7 bits, 11 bits and 9 bits respectively while using the variable length fixed point number system. This reduces memory utilization by 28.6\%, 36.4\% and 23\% as compared to standard state of the art variable length fixed point implementations\cite{judd2015reduced,judd2017proteus11}, and can also significantly reduce the number of memory accesses through memory concatenation schemes. Note that this improvement is achieved without using quantization or retraining the DCNNs.

\section{Conclusion}
We explore using the posit number system to represent weights of three DCNNs on MNIST, Cifar-10, and ImageNet datasets. The normalized posit number system outperformed the fixed point number system in terms of accuracy, with fewer number of bits used to represent weights. By using a memory concatenation encoding scheme, the number of memory accesses required to transfer the weights reduces significantly, as well as the total energy consumption for the same task. For future work, we will explore the effect of low precision data representation using the posit number system and implement DCNNs using the posit representation for both storage and computation.

\bibliographystyle{IEEEtran}
\bibliography{ACM}

\end{document}